\def\year{2022}\relax
\documentclass[letterpaper]{article} 
\usepackage{aaai22}  
\usepackage{times}  
\usepackage{helvet}  
\usepackage{courier}  
\usepackage[hyphens]{url}  
\usepackage{graphicx} 
\usepackage{natbib}  
\usepackage{caption} 
\DeclareCaptionStyle{ruled}{labelfont=normalfont,labelsep=colon,strut=off} 
\usepackage{algorithm}
\usepackage{algorithmic}
\usepackage{tikz}
\newcommand*\circled[1]{\tikz[baseline=(char.base)]{
            \node[shape=circle,draw,inner sep=0.8pt] (char) {#1};}}
\usepackage{newfloat}
\usepackage{listings}
\floatstyle{ruled}
\newfloat{listing}{tb}{lst}{}
\floatname{listing}{Listing}
\title{Design of Explainability Module with Experts in the Loop for Visualization and Dynamic Adjustment of Continual Learning}
\author{
    Yujiang He\thanks{Accepted at the AAAI-22 Workshop on Interactive Machine Learning (IML@AAAI'22)},
    Zhixin Huang,
    Bernhard Sick
}
\affiliations{
    Intelligent Embedded Systems Lab, University of Kassel\\

    Wilhelmshöher Allee 67\\
    34121, Kassel\\
    yujiang.he$|$zhixin.huang$|$bsick@uni-kassel.de
}

\begin{document}

\maketitle

\begin{abstract}
Continual learning can enable neural networks to evolve by learning new tasks sequentially in task-changing scenarios.
However, two general and related challenges should be overcome in further research before we apply this technique to real-world applications.
Firstly, newly collected novelties from the data stream in applications could contain anomalies that are meaningless for continual learning.
Instead of viewing them as a new task for updating, we have to filter out such anomalies to reduce the disturbance of extremely high-entropy data for the progression of convergence.
Secondly, fewer efforts have been put into research regarding the explainability of continual learning, which leads to a lack of transparency and credibility of the updated neural networks.
Elaborated explanations about the process and result of continual learning can help experts in judgment and making decisions.
Therefore, we propose the conceptual design of an explainability module with experts in the loop based on techniques, such as dimension reduction, visualization, and evaluation strategies.
This work aims to overcome the mentioned challenges by sufficiently explaining and visualizing the identified anomalies and the updated neural network.
With the help of this module, experts can be more confident in decision-making regarding anomaly filtering, dynamic adjustment of hyperparameters, data backup, etc.
\end{abstract}

\section{Introduction}
\label{sec:intro}
Continual learning enables neural networks to accumulate novel knowledge for handling unknown tasks sequentially through learning in task-changing contexts, where distributions of the source domain and target domain can change over time.
It aims to overcome the so-called catastrophic forgetting problem~\cite{mccloskey1989catastrophic,ratcliff1990connectionist}, i.e., a trained neural network could forget the obtained knowledge regarding previous tasks after learning to handle novel tasks.
Numerous continual learning algorithms have been proposed and evaluated in experiments with public datasets, such as~\cite{kirkpatrick2017overcoming,zenke2017continual,li2017learning,chaudhry2018riemannian}.

Novelties refer to the samples that a pre-trained neural network has never seen.
The bulk of novel samples that appear in applications could result in concept drift, which leads to a decrease in the performance of the neural network.
In the standard experimental setup of continual learning for regression tasks, all novelties can be viewed as new tasks for updating the model using continual learning~\cite{he2021clear,he2021toward}.
The goal is to maintain the performance of the neural network at a high level throughout its lifecycle.
Generally, neural networks learn multiple tasks sequentially with fixed learning hyperparameters, usually searched beforehand by intuition or expert prior knowledge.
Subsequently, the updated model will be evaluated on a validation dataset containing known tasks.
Standard evaluation metrics for continual learning are the same as in the traditional deep learning experiments, e.g., accuracy and precision, which are understandable but barely comprehensive.

Compared with the performance of the continual learning algorithms in experiments, an unexpected decrease of the performance could be caused in real-world applications because of the following shortcomings in the general setup:
\begin{itemize}
    \item{\textbf{Confusing anomalous and novel examples}}. 
    The samples that the current neural network can not predict accurately can be defined as novelties. The network should continually learn to update its knowledge memory as data is accumulated.
    However, anomalies, i.e., samples that are meaningless for continual learning, also belong to the novelty group.
    Anomalies are usually caused by unforeseen incidents, such as damaged sensors and stopped energy generators under specific conditions.
    Learning meaningful novelties can instill a neural network with new abilities to handle non-stationary data streams. 
    Still, learning anomalies without restriction could mislead the neural network to a wrong converged state and decrease its performance.
    \item{\textbf{Fixed hyperparameters}}.
    Learning all tasks with the fixed hyperparameters indicates that the algorithm must be robust to the chosen hyperparameters.
    Hyperparameters choices are usually tuned manually by prior knowledge and intuition or searched by optimization methods, such as grid search or random search, based on the currently available datasets. 
    However, the constant hyperparameters could be sensitive to unknown tasks and then disturb the evolution of the neural network in the long run.
    \item{\textbf{A lack of comprehensive evaluation metrics}}.
    Standard evaluation metrics, such as accuracy and F1-score for classification problems, can reflect only the performance of neural networks regarding the given dataset at a specific time point, i.e., at the convergence after the last training epoch.
    We think a set of comprehensive metrics should evaluate the updating results from various perspectives besides accuracy.
    For example, calculating the extent to which catastrophic forgetting by quantifying the overlap of the distributed representation~\cite{french1992semi}, or counting the free neurons to estimate the remaining free space of the neural network, which determines whether subsequent tasks can be learned continually and successfully~\cite{mirzadeh2020dropout}.
    A set of comprehensive metrics can timely alarm supervisors to react to an unsatisfied update.
    \item{\textbf{Assessment with access to old datasets}}.
    In some cases, dependency on old datasets for learning and evaluation might violate some regulations, such as the General Data Protection Regulation, which the European Union introduced to remedy potential data privacy problems.
    The regulation states that the data subject has the right to request the erasure of personal data, which requires learning and evaluation without access to the old dataset under some circumstances. 
    \item{\textbf{A lack of visualization methods}}.
    Numeric evaluation can give us only an abstract score for detected anomalies or updated neural networks.
    A lack of elaborated visualization leads to researchers' uncertainty in the identified anomalies or the updated models.
    Furthermore, despite being possible to analyze the generalization performance of neural networks by observing the downward trend of the losses over training epochs, the progress regarding the dynamic evolution of neural networks' hidden layers is still opaque.
    Humans can not understand and interpret complex, high-dimensional data without proper visualization methods.
\end{itemize}

In this article, we propose thus a conceptual design of an explainability module with experts in the loop for addressing the problems that are mentioned above and generally exist in current applications of continual learning.
The goal of the explainability module is to provide sufficient explanation regarding the identified anomalies and details of model updating to experts.
Based on the interaction between the explainability module and experts, we expect to achieve real-time monitoring and dynamic management for neural networks with continual learning.

The remainder of this article starts with a description of our conceptual design, which integrates and extends the existing techniques in terms of evaluation and visualization.
Then in Sec.~\ref{sec:usecase_outlook}, we present a specific use case about how the proposed design can serve as an extension to the existing work in~\cite{he2021clear}, an adaptive continual learning framework for solving regression problems in non-stationary contexts.
The article is concluded with a brief outlook of our further research.
\section{Design of Explainability Module}
\begin{figure*}[htp]
\centering
\includegraphics[width=1.92\columnwidth]{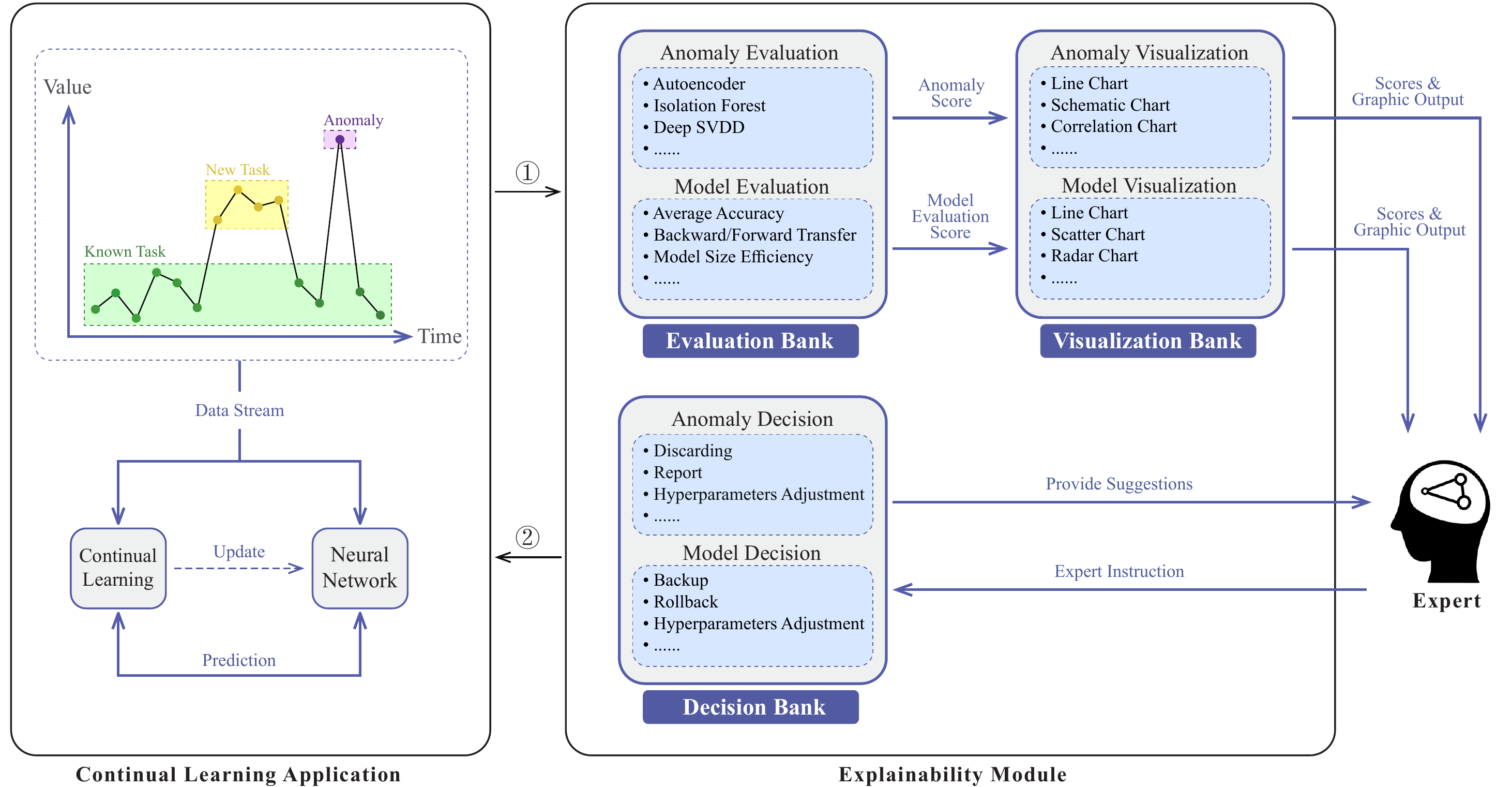} 
\caption{An illustration about how the explainability module with experts in the loop manages the data stream and the neural network dynamically in a general continual learning application for regression. Based on the data in the interaction information flow~\circled{1}, the evaluation bank can compute assessment scores for detected anomalies or the neural network updated with the new task and then send the data as well as the scores to the following bank for visualization. The scores and graphic outputs can offer experts insight into the details of the data and the model. Sequentially, experts can easily make an accurate decision and then feed it back via the information flow~\circled{2}.}
\label{fig:X_module}
\end{figure*}

Figure~\ref{fig:X_module} illustrates the interaction between a continual learning application and the proposed explainability module with an expert in the loop.
A new task, also called as an unknown task, is defined differently in the research community of continual learning.
For example,~\cite{lomonaco2017core50} propose three types of new tasks for object recognition, which are new instances (NI), new classes (NC), and new instances and classes (NIC), respectively.
NI and NC refer to new training patterns of the same classes and new training patterns belonging to the different classes, correspondingly.
NIC is a combination of NI and NC, i.e., the new training patterns belonging to known and new classes.
In comparison, a new task in regression is hard to define due to a lack of labels.
\cite{he2021clear} consider a new regression task as a group of samples that decrease the predictive accuracy of the current neural network.
In Fig.~\ref{fig:X_module}, we display an example of a data stream containing a known regression task, a new regression task, and an anomaly.

The neural network can predict the known task and evolve with the new task by continual learning.
The identified anomaly will be transferred to the explainability module via the interaction information flow~\circled{1} for a precise assessment.
Besides the anomalies, the flow~\circled{1} also consists of other application-related details, including the samples from the data stream, the corresponding predictions, hyperparameters of learning, the updated neural network, an elaborated updating log, etc.
The interaction can happen once an anomaly is detected or an update is done.

As shown in Fig~\ref{fig:X_module}, the explainability module contains three primary submodules: evaluation bank, visualization bank, and decision bank, each of which has two blocks.
The anomaly block is designed to analyze the potential anomalies identified in the data stream, and the model block can evaluate the updated neural network.
Experts can observe and judge whether a positive-true anomaly exists in the data stream and decide how to process it.
Besides, experts can decide to keep a backup of a successfully updated model or restore a failed updated model with an adjustment of the hyperparameters after assessing the updating quality based on the evaluation metrics and visualization.
The experts are the final decision maker with professional knowledge and experience.
The main function of our explainability module is to provide suggestions and explanations to help the experts to make decisions.
The final decisions regarding the anomalies and the updated models are transferred via the interaction information flow~\circled{2}.

The remainder of this section elaborates the three submodules of our design and introduces the existing techniques relevant to implementing the explainability module.

\subsection{Evaluation Bank}
\label{subsec:evaluation_bank}
\subsubsection{Anomaly Evaluation}
The anomaly evaluation block is responsible for evaluating the potential anomalies by computing an anomaly score, which assists experts in decision-making.

In a traditional deep learning setup, training and validation datasets are preprocessed before experiments by discarding anomalies to eliminate the negative influence of insufficient data and efficiently improve the training results.
However, anomalies in real-world applications are unavoidable, bringing high but meaningless entropy to the training process and finally disturbing the updating results.
Therefore, retaining meaningful samples and filtering out anomalies from novelties plays a crucial role in maintaining neural networks' performance in continual learning applications.

Lots of machine learning-based anomaly detection methods have been proposed, including supervised and unsupervised methods.
Deep Support Vector Data Description (Deep SVDD) maps samples from the input space to the output space using a neural network. 
It identifies anomalies due to the distance between the samples and centers of clusters~\cite{ruff2018deep}.
Samples that are far away from the cluster center of normal samples are marked as anomalies.
\cite{ding2013anomaly} propose an unsupervised method for anomaly detection, called the Isolation Forest Algorithm for the Stream Data (IFASD).
It is a variety of Isolation Forest focusing on multidimensional time series.
Besides,~\cite{transformer_recon} present a transformer-based unsupervised method to detect anomaly by comparing the error between the original time series and the reconstructed time series.
These methods evaluate the quality of the detected anomalies by calculating a score and comparing the score with a predefined threshold.
If the score is above the threshold, the corresponding novelties are marked as anomalies rather than a new task for continual learning.

For supervised methods, samples must be labeled manually, which is time-consuming and expensive in real-world applications~\cite{ref_AL_AD}.
By comparison, using unsupervised methods can avoid additional manual labels and computation overhead. 
Still, trained models might have a very high false-positive rate, requiring further expert verification of the identified anomalies~\cite{ref_AL_Difficulties}.

\subsubsection{Model Evaluation}
The model evaluation can assess the quality of the updated model by providing an evaluation score based on a comprehensive metric.

We consider that a comprehensive evaluation metric should consider not only the improved prediction accuracy for new tasks but also the dropping accuracy for old tasks, also known as forgetting, during continual learning.
A trade-off between obtaining knowledge from new tasks and retaining knowledge from old tasks, the so-called plasticity-stability dilemma~\cite{mermillod2013stability}, is dependent on the specific application scenario.
For example, in the application of renewable energy forecasting, the aging of power generators is a slow-changing and irreversible process.
In this case, more attention can be paid to how neural networks can fit new data well and slightly discard the stored memory.
However, in the application of meteorological time series forecasting, weather conditions change seasonally, where the probability distribution of data shifts periodically over time.
Memory regarding historical data should be consolidated while neural networks learn new data.

\cite{lopez2017gradient} define three metrics for evaluating continual learning experiments: Average Accuracy, Backward Transfer, and Forward Transfer.
The goal of the metrics is to assess the influence of learning a new task on the performance of the trained network regarding all known tasks, including previous and future tasks.
\cite{2018dont} extend this work by proposing additional three metrics, i.e., Model Size Efficiency, Sample Storage Size Efficiency, and Computational Efficiency, which measure the increment of model memory size, storage overhead of samples for replay-based strategies, and computational overhead for learning all tasks.
For ranking purposes, these metrics are summed to a continual learning score with different weighting schemes.
Besides, ~\cite{he2020continuous,he2021clear} evaluate continual learning strategies in the context of regression problems in terms of prediction error, fitting error, forgetting ratio, and training time.
The idea behind is similar to~\cite{2018dont}.

\subsection{Visualization Bank}
\label{subsec:visualization}
\subsubsection{Anomaly Visualization}
We use the anomaly visualization to generate the visualization outputs of these anomalies, which can explain the abstract anomaly score in a trustable and understandable way.

High-dimensional data usually indicates multiple data sources, e.g., multiple sensors integrated into a complex system, which brings us two main challenges for anomaly detection. 
Firstly, it is difficult to manually analyze and visualize high-dimensional data without losing local and global architecture information.
Secondly, some data sources are dependent, which means that one faulty source could affect other normal ones.
The dependency makes it too difficult to identify the exact faulty source and when the fault occurs.
Our proposed anomaly visualization module aims to solve these challenges by figuring out the following questions:
\begin{itemize}
    \item How to utilize dimension reduction techniques to make the visualization of the high-dimensional data understandable?
    \item How to visualize the dependency among multiple data sources for providing reasonable explanations for detected anomalies? 
    \item How to rank anomaly detection methods if available methods or metrics are multiple?
\end{itemize}

\subsubsection{Model Visualization}
The model visualization focuses on visualizing the dynamic evolving process of the updated model, including the change of model parameters from the random initialization to convergence.
Compared to the learning curve, where each point represents the prediction error at each time point, we can be more convinced by the visualization regarding the dynamic evolution during continual learning, which can show more details of the progress.
For achieving this goal, we need to consider how to preserve the information of the data as much as possible in the process of dimension reduction.

Common dimension reduction techniques, such as Principal Component Analysis (PCA), diffusion maps~\cite{coifman2006diffusion}, multidimensional scaling (MDS)~\cite{cox2008multidimensional}, t-SNE~\cite{vandermaaten08a}, can efficiently preserve local or global structure information.
\cite{moon2019visualizing} propose PHATE, a method that can visualize high-dimensional data while revealing both local and global structures.
They compare PHATE and other methods on biological and artificial datasets.
Based on PHATE,~\cite{gigante2019mphate} propose M-PHATE, which visualizes dynamic evolution of neural networks by extracting changes of hidden representations over epochs.
Besides,~\cite{mirzadeh2020dropout} selectively visualize active neurons in each task using a heat map, which is intuitive but not applicable to giant neural networks.

Similar to the anomaly visualization, we also need to rank continual learning algorithms under various evaluation metrics for the model visualization, for example, to compare different algorithms or neural network architectures in ensemble learning.
Common visualization methods, such as radar charts and line charts~\cite{2018dont}, are intuitive and easily implemented for ranking purposes.
It makes the updated model more transparent and trustworthy.

The inputs of the visualization bank are the interaction information flow~\circled{1} and the output of the evaluation bank. 
Their graphic outputs provide experts insights into latent representations of data and models and support experts in making decisions.
Furthermore, we expect that the graphic outputs are interactive rather than static.
They can, for example, zoom in/out, display the exact values, and easily be exported and saved in a readable format.

\subsection{Decision Bank}
\label{subsec:decisionmaker}
\subsubsection{Anomaly Decision}
The scores and graphs of anomalies from the first two banks are sent to experts for manual analysis.
According to~\cite{real_work_anomaly}, novelties can be grouped into four classes: sensor failure, component failure, system failure, and concept drift.
Decision bank could offer experts these suggestions for the categories of the novelties, as well as the corresponding decisions:
\begin{itemize}
    \item In the context of sensor failure, where only one of the sensors behaves abnormally, we can replace the data from the fault sensor with the data from the highly relevant sensor, rather than discarding all data of these sensors.
    \item Component failure denotes that multiple low-correlated sensors in a component behave abnormally at the same time.
    Low correlation means that these sensors may not have a cascade relationship, thus indicating the defect of the component.
    The whole system might break down if the component is connected with others.
    Such novelties should be stored in an anomaly database for further analysis and physical system maintenance rather than updating.
    \item System failure is similar to multi-component failure, where faults are detected in multiple components.
    A system failure could be caused by multiple faulty components or a chain reaction caused by a single faulty component.
    Like component failure, such a system failure should also be stored in the anomaly database instead of being learned.
    Besides, manual intervention could also lead to system failure.
    For example, wind turbines have to be stopped by a shut-off mechanism under the condition of high wind speed to avoid significant damage.
    If the manual intervention is long-term and periodical, such a system failure could be learned by neural networks.
    \item Concept drift means that the error between predictions of neural networks and measurements of sensors starts to increase from some time point due to change of source or target data distribution.
    It often occurs when, for example, sensors or devices are replaced, sensor parameters are adjusted, or environmental and temporal changes.
    This kind of novelty should be viewed as a new task for updating.
\end{itemize}

According to the analysis and suggestions given by the module, experts can decide to discard these anomalies or use them for updating models.
In the case of too many misclassified novelties in anomaly evaluation, experts can adjust the related parameters, such as the threshold, or update the anomaly detector to reduce the false-positive rate.

\subsubsection{Model Decision}
There are three significant decisions for the updated neural network: backup, rollback, and hyperparameters adjustment.
These decisions are relevant.

The backup means storing the neural networks and the datasets after a successful update.
The rollback denotes that the updated model returns to the backup state as an initialization state for a new update if the update is failed.

Here, a successful update is defined as the loss function gradually converges to a low state in learning unknown tasks without a noticeable decrease of performance of neural networks on old tasks.
Parameters of the successfully updated model should be stored as a backup.
According to the types of replay-based continual learning algorithms, data backup can denote storing training datasets regarding known tasks in an original format or training a generative adversarial network with current tasks to generate fake data.

By contrast, a failed update means that the loss can not be converged within the given training epochs or the updated model loses most knowledge regarding old tasks in the evaluation phase.
The process of optimization using common continual learning algorithms is minimizing a target loss function with stochastic gradient descent.
Therefore, stochasticity in the training process could cause a failed update, which leads to catastrophic forgetting.

In addition, the improper hyperparameters or the undetected anomalies in the training dataset can also lead to a failed update.
Therefore, the decision bank can display all adjustable hyperparameters to experts and provide explicit suggestions with corresponding explanations.
Experts can adjust the hyperparameters by their intuitions and professional experience for the future application or a new round of updating under the case where the  updating result is unacceptable.
\section{Use Case \& Outlook}
\label{sec:usecase_outlook}
\subsection{Use Case: The CLeaR Framework}
CLeaR, short for continual learning for regression problems, is an adaptive framework for overcoming the catastrophic forgetting problem in a non-stationary regression application scenario~\cite{he2021clear}.
Samples in the real-time data stream are categorized into novelties and familiarities.
Novelties are the samples that the trained neural network can not predict accurately, while familiarities are the samples that the network is familiar with.
CLeaR distinguishes the two types by computing the error between the truth and the prediction and comparing the error with a threshold, which is dynamically adjustable according to the network's convergence after training.
Novelties and familiarities are stored in two independent buffers for continual learning.
Updating can be triggered under customizable conditions.
For example,~\cite{he2021clear} design a novelty buffer with a fixed length and trigger updating when the novelty buffer is full.

The five problems mentioned in Sec.~\ref{sec:intro} exist in the current design of CLeaR.
Firstly, the boundary between the novelty and anomaly is ambiguous according to the categorization, possibly leading to a failed update because of misclassification.
Secondly, the experimental setups of CLeaR in~\cite{he2021clear} and~\cite{he2021toward} are the same, i.e., learning all detected unknown tasks with a set of fixed parameters that are optimized beforehand.
Thirdly, the influences of two hyperparameters of the CLeaR framework, i.e., the novelty buffer size and the threshold factor, on its learning performance have been analyzed in~\cite{he2021clear}.
The two parameters control the frequency of triggered updates and adjust the judgment mechanism for novelty detection, respectively, thus balancing the performance of the updated model on all tasks and affecting the computational overhead in applications.
Nevertheless, a comprehensive evaluation of the robustness of these hyperparameters is still missing in the previous research.
Because these components in the framework, such as the type of the neural network and continual learning algorithms, the size of the buffers, and the novelty detection method, are customizable depending on the requirements of the specific application.
These hyperparameters should be adjusted dynamically during the application in order to learn different tasks efficiently.
Besides, the proposed module can improve the explainability and transparency of the CLeaR framework by providing visualization and comprehensive evaluations.

\subsection{Outlook}
This paper analyzes the problems and shortcomings that can appear when continual learning is applied to real-world scenarios that have the same setup as in experiments.
For solving these problems, we propose the conceptual design of the explainability module integrated with various deep learning-related techniques, such as anomaly detection, evaluation, and visualization.
In our further research, we plan to implement the design and apply it to the existing the CLeaR framework as an extension.
Based on the visual and explainable outputs of the module, experts can dynamically manage the evolution of neural networks driven by continual learning, and perform more reliable evaluation to its updates in task-changing context.
\section*{Acknowledgments}
This work is supported within the Digital-Twin-Solar (03EI6024E) project, funded by BMWi: Deutsches Bundesministerium für Wirtschaft und Energie/German Federal Ministry for Economic Affairs and Energy.
We also want to thank our colleagues in the Intelligent Embedded Systems Lab at the University of Kassel, Kristina Dingel and Lukas Rauch, for their constructive reviews and suggestions.

\bibliography{aaai22}
\end{document}